\documentclass[preprint,12pt]{elsarticle}

\usepackage{times}
\usepackage{epsfig}
\usepackage{graphicx}
\usepackage{amsmath}
\usepackage{amssymb}
\usepackage{multirow}
\usepackage{algorithm}
\usepackage{algpseudocode}
\usepackage{subcaption}
\usepackage[T1]{fontenc}
\usepackage{hyperref}
\usepackage{tabularx}
\usepackage{arydshln}
\usepackage{bbding}

\journal{Engineering Applications of Artificial Intelligence}

\begin{document}

\begin{frontmatter}



\title{Balance Divergence for Knowledge Distillation}



\author[1]{Yafei Qi}
\ead{qiyafei@csu.edu.cn}
\affiliation[1]{School of Computer Science and Engineering, Central South University, Changsha, 410073, Hunan, China}

\author[2]{Chen Wang}
\affiliation[2]{Institute of Artificial Intelligence, Shaoxing University, Shaoxing, 312000, Zhejiang, China}

\author[3]{Zhaoning Zhang}
\affiliation[3]{Computer Science and Technology, National University of Defense Technology, Changsha, 410073, Hunan, China}

\author[4]{Yaping Liu}
\affiliation[4]{Cyberspace Institute of Advanced Technology, Guangzhou University, Guangzhou, 510006, Guangdong, China}

\author[1]{Yongmin Zhang\corref{cor1}}
\ead{zhangyongmin@csu.edu.cn}

\begin{abstract}
Knowledge distillation has been widely adopted in computer vision task processing, since it can effectively enhance the performance of lightweight student networks by leveraging the knowledge transferred from cumbersome teacher networks. 
Most existing knowledge distillation methods utilize Kullback-Leibler divergence to mimic the logit output probabilities between the teacher network and the student network.
Nonetheless, these methods may neglect the negative parts of the teacher's ``dark knowledge'' because the divergence calculations may ignore the effect of the minute probabilities from the teacher's logit output.
This deficiency may lead to suboptimal performance in logit mimicry during the distillation process and result in an imbalance of information acquired by the student network. 
In this paper, we investigate the impact of this imbalance and propose a novel method, named Balance Divergence Distillation. 
By introducing a compensatory operation using reverse Kullback-Leibler divergence, our method can improve the modeling of the extremely small values in the negative from the teacher and preserve the learning capacity for the positive. Furthermore, we test the impact of different temperature coefficients adjustments, which may conducted to further balance for knowledge transferring.
We evaluate the proposed method on several computer vision tasks, including image classification and semantic segmentation. 
The evaluation results show that our method achieves an accuracy improvement of $1\%\sim3\%$ for lightweight students on both CIFAR-100 and ImageNet dataset, and a $4.55\%$ improvement in mIoU for PSP-ResNet18 on the Cityscapes dataset. The experiments show that our method is a simple yet highly effective solution that can be smoothly applied to different knowledge distillation methods. 
\end{abstract}


\end{frontmatter}


\section{Introduction}
In recent years, deep learning convolutional neural networks have been widely applied to various computer vision tasks, such as classification, semantic segmentation, and object detection. 
Generally the models with more computational complexity and structural complexity often have better results. However, larger models also bring challenges concerning lower efficiency and difficulties in deploying them to devices with limited computational resources. To address this problem, researchers have proposed a series of model compression methods, including model pruning \cite{DBLP:journals/corr/HanMD15,liu2017learning}, model quantization \cite{courbariaux2016binarized,DBLP:conf/iclr/0022KDSG17}, and knowledge distillation (KD) \cite{hinton2015distilling,DBLP:journals/corr/RomeroBKCGB14}. Among these, KD has gained widespread use across various computer vision applications.

\begin{figure}[t]
   \centering
      \begin{subfigure}[b]{\textwidth}
         \centering
         \includegraphics[width=0.8\textwidth]{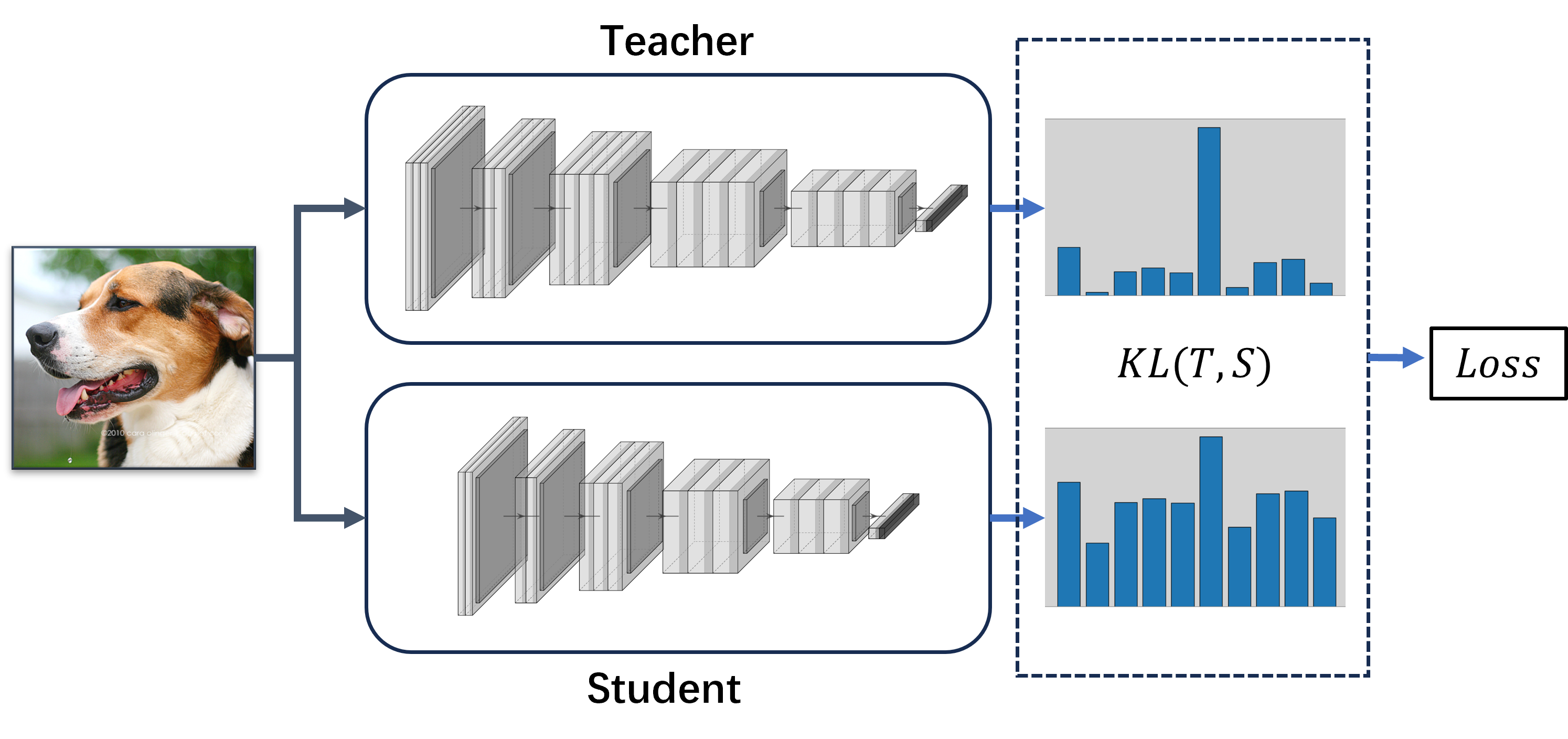}
         \caption{Output probability fitting by normal KD.}
         \label{fig:fig0a}
      \end{subfigure}
      
      \begin{subfigure}[b]{\textwidth}
         \centering
         \includegraphics[width=0.8\textwidth]{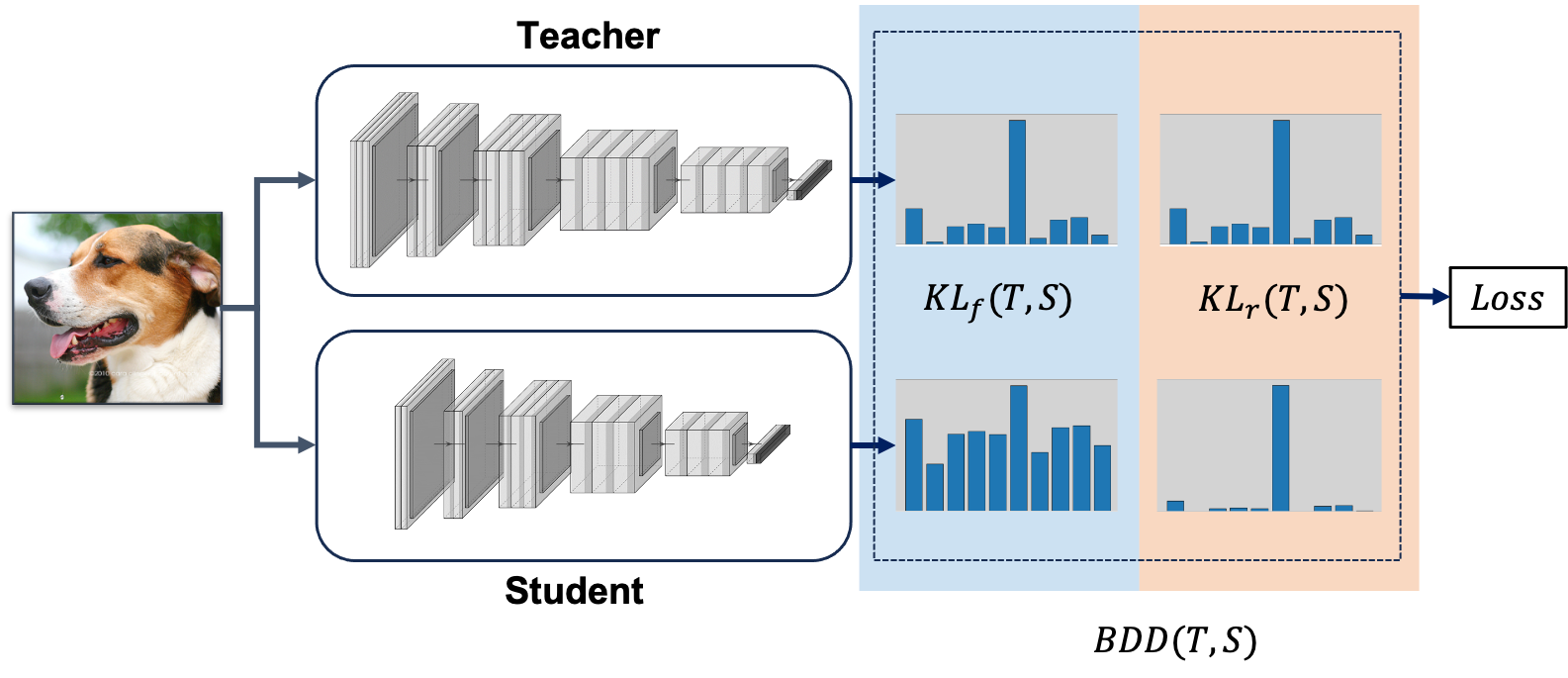}
         \caption{Output probability fitting by BDD.}
         \label{fig:fig0b}
      \end{subfigure}
      
      \caption{\textbf{Comparison of output probability fitting by normal KD and BDD}. As shown in figure a is normal KD, the minuscule values can not learn well by student. Figure b is our proposed Balance Divergence Distillation(BDD), we highlight the positive and negative regions of the teacher's probability outputs through temperature coefficient scaling, and then use different KL divergences in BDD to mimic the model's outputs.} 
   \label{fig:fig0}
\end{figure}
KD was first introduced by Hinton \cite{hinton2015distilling}, primarily to minimise the disparity between the logit outputs of cumbersome teacher networks and lightweight student networks by using Kullback-Leibler (KL) divergence. As KD methods progress, feature-based\cite{liu2020adaptive,yang2022masked} and relation-based\cite{chen2021distilling,shu2021channel} KD methods have also been introduced alongside logit-based distillation. Feature-based distillation aims to reduce differences in feature maps between teacher and student, while relation-based distillation aims to minimize the inter-feature relationships between the teacher and student. These approaches offer additional strategies to enhance different forms of knowledge transfer in KD. Commonly, the methods using feature or releation knowledge typically involve additional computational costs to align the feature maps or to discover mutual relationships between teacher and student. To enhance the employment of logit-based distillation methods, some works\cite{li2023curriculum,zhao2022decoupled,Jin_2023_CVPR} have started to optimize from the perspective of balancing the positive and negative parts of the logit-based representation. However, most of them neglect the basic imbalance in the loss calculation of KL divergence. 

In this paper, we aimed to explore the potential limitations of the loss used in logit-based KD methods and conducted an in-depth investigation of the loss balance methodology.
We studied the KL divergence calculation process and found that the KL divergence calculation tends to focus more on the higher probability outputs (positive samples) of the teacher, while neglecting the calculation of extremely small values (negative samples). 
This potential bias may hinder the ability of the student network to accurately capture all of the teacher's output information. 
Therefore, we proposed an innovative method called Balanced Divergence Distillation (BDD), which improves the balance in the calculation of KL divergence by introducing reverse KL divergence and adjusting the temperature coefficient. 
Figure (\ref{fig:fig0}) demonstrates the computational process of the BDD technique. When conducting the computation, the KL divergence was separated into two categories, namely forward-KL and reverse-KL, with different components emphasising the different positive and negative values.
At the same time, BDD goes a step further in balancing the calculation of different KL divergences by adapting the supervised information from the teacher's outputs by adjusting the temperature, enabling the student to gain more comprehensive knowledge from the teacher.
Our work demonstrates that a simple modification of the KL divergence calculation may effectively address the problem of imbalance and achieve promising results across different tasks. Our main contributions are as follows:

\begin{itemize}
   \item We conducted an in-depth investigation and demonstrated the imbalance problem present in KL divergence, and subsequently proposed the BDD method, which distinguishes between forward and reverse KL divergence, effectively addressing the imbalance problem.
   \item We further explored the coordination mechanism of temperature coefficients between forward and reverse KL divergences, and investigated improvements in different attentional outputs that may enhance the ability to address the imbalance problem.
   \item We verified the efficacy of BDD in both classification and dense prediction tasks. The experiments revealed an accuracy improvement ranging $1\%\sim3\%$  for lightweight networks on the CIFAR-100 and ImageNet datasets, and a notable $4.55\%$ enhancement in mIoU for PSP-ResNet18 when evaluated on the Cityscapes dataset. These results demonstrate that BDD can easily integrate with other methods and achieve promising results across different tasks.
\end{itemize}

The paper is organized as follows: Section 2 offers a brief overview of related work in knowledge distillation, Section 3 provides a comprehensive description of the proposed BDD methodology, Section 4 presents and analyzes the experimental results, and finally, Section 5 concludes the paper.

\section{Related Work}
The primary objective of knowledge distillation methods is to address the problem of overly hard label supervision\cite{hinton2015distilling}. Since the teacher network demonstrates superior generalisation abilities, KL divergence is widely used to provide the student with softer labels in knowledge distillation methods. The loss of KL divergence often referred to as ``dark knowledge'' from the teacher, leading to performance improvements. Subsequent work can be categorized into three main directions: logit-based \cite{cho2019efficacy,li2023curriculum,DBLP:conf/aaai/MirzadehFLLMG20,DBLP:conf/iclr/ZhouSCZWYZ21,zhao2022decoupled}, feature-based \cite{heo2019comprehensive,heo2019knowledge,liu2020adaptive,DBLP:journals/corr/RomeroBKCGB14,yang2022masked}, and relation-based \cite{chen2021distilling,park2019relational,shu2021channel,DBLP:conf/iclr/TianKI20,wang2020intra} approaches.

\subsection{Logit-based Knowledge Distillation}
Logit-based methods mainly focus on improving the ability of the teacher to transfer balance information from soft labels.
Cho \cite{cho2019efficacy} found that larger networks are not always better teachers, as significant model differences may prevent students from effectively imitating teacher performance. 
TAKD \cite{DBLP:conf/aaai/MirzadehFLLMG20} addresses the disparities between large teacher networks and lightweight student networks by introducing an intermediate scale teacher assistant that bridges the significant gap between teacher and student, which make knowledge transferred between the models more balanced. 
WSL \cite{DBLP:conf/iclr/ZhouSCZWYZ21} considers the different bias-variance effects brought about by soft-label distillation from the perspective of regularization samples and proposes a weighted soft label approach, which results in a more balanced learning of information. 
DKD \cite{zhao2022decoupled} introduced a balanced decoupled knowledge distillation method, which utilized label signals for supervision and distilled the target class attention and non-target class attention between teacher and student networks. 
BD-KD \cite{amara2022bd} introduced symmetric KL divergence in online knowledge distillation, allowing both the teacher and student to learn more effectively. 
CTKD \cite{li2023curriculum} considers the effect of temperature coefficient on knowledge distillation and introduces a dynamically learnable temperature coefficient adjustment method constructed by reverse gradients to balance the learning process of the student network. 
MLD \cite{Jin_2023_CVPR} aligns predictions at the instance, batch, and class levels to balance and incorporate the teacher's knowledge comprehensively.

\subsection{Feature-based and Relation-based Knowledge Distillation}
Currently, many state-of-the-art (SOTA) methods primarily fall into the categories of feature-based and relation-based approaches. 
OFD \cite{heo2019comprehensive} and AB \cite{heo2019knowledge} argue that activation layers in neural networks possess more extensive feature knowledge and improve distillation performance by exploring activation boundaries.
RKD \cite{park2019relational} mainly focuses on relationships between data samples, guiding the learning of student by minimizing the distance and angle between outputs from different data samples. 
CRD \cite{DBLP:conf/iclr/TianKI20} originates from the structured knowledge of teacher networks and proposes a contrastive loss for distilling student networks. 
AMTML \cite{liu2020adaptive} uses multiple teacher networks to generate adaptively integrated soft labels and intermediate layers to generate balance information for student network distillation. 
ReviewKD \cite{chen2021distilling} improves student distillation performance by constructing knowledge through reviewing the outputs of different feature layers in the teacher network. 
IFVD \cite{wang2020intra} focuses on the intra-class relationships of pixels with the same label information, transferring knowledge by minimizing the cosine distance of intra-class relationship. 
Attention \cite{DBLP:conf/iclr/ZagoruykoK17, guo2023class, li2023method, liu2024image} plays a crucial role in knowledge distillation, which often involves defining attention maps appropriately and forces student to mimic the attention maps of teacher network.
Channel-wise distillation methods \cite{shu2021channel,gou2022channel} change pixel-wise knowledge transfer to channel-wise probability outputs in dense prediction tasks to ensure the balance of information within the same channel. 
Masked generative distillation \cite{yang2022masked, jiang2023masked, lao2023masked} enhances the performance of the student network by randomly masking student feature information, and forcing it to learn complete features from the teacher network.


Additionally, there exist domain-specific knowledge distillation methods that are not universally applicable. For instance, in object detection\cite{zheng2022localization,cao2022pkd,li2023progressive,tang2023task}, methods concentrate primarily on enhancing the transfer of task-specific foreground and background knowledge or different spatial knowledge, which could also be considered as a kind of information balance during the distillation process. 
Usually, logit-based methods need more supervision information and may not achieve optimal performance. On the other hand, feature-based, relation-based and domain-specific methods can achieve better performance, but often add complexity and computational cost.
However, These methods rarely account for the unbalanced information passed on by the basic KD loss of KL divergence. This paper primarily concentrates on a well-balanced divergence utilised in a logit-based knowledge distillation method.

\section{Proposed Method}
\subsection{Preliminary}
Firstly, we introduce the notations commonly used in logit-based knowledge distillation (KD) frameworks.

\textbf{Logit Output.} Logit output also known as a probability output, can be attained from the prediction output of a network. In classification tasks, the objective is to predict the correct class among the $C$ categories in the output layer $F_{1 \times 1 \times C}$. In dense prediction tasks like semantic segmentation, the goal is to assign each pixel to one of the $C$ categories, and the logit output is denoted by $F_{H \times W \times C}$, where $H$ and $W$ specify the height and width of the output layer, respectively.
Generally, let $F \in \mathbb{R}^{H \times W \times C}$ and $Y \in \mathbb{R}^{H \times W \times 1}$ represent the output of the network and ground truth label.
The softmax function $p$ is often used to normalize output into logit output applied on the output layer $F$, and the cross-entropy loss $\mathcal{L}_{CE}$ is commonly used to reduce the loss between logit output and ground truth label during training, which can be defined as:
\begin{equation}
   \begin{aligned}
      \mathcal{L}_{CE} &= -\frac{1}{H \times W}\sum_{i=1}^{H \times W}\sum_{c=1}^{C}Y_{i,c}\log{p(F_{i,c})}\\
      & p(F_{i,c}) = \frac{e^{F_{i,c}}}{\sum_{c=1}^{C}e^{F_{i,c}}}
   \end{aligned}
\end{equation}

\textbf{KL-Divergence.} Knowledge Distillation (KD) methods commonly use functions such as Kullback-Leibler (KL) divergence \cite{hinton2015distilling} or mean square error (MSE) \cite{DBLP:journals/corr/RomeroBKCGB14} as loss to transfer knowledge from a teacher network to a student. In accordance with Hinton's work \cite{hinton2015distilling}, we formulate KL divergence as the loss function in our knowledge distillation framework:

\begin{equation}\label{eq2}
   \mathcal{KL}(p(F^T) || p(F^S)) = \frac{1}{H \times W}\sum_{i=1}^{H \times W}\sum_{c=1}^{C}p(\frac{F_{i,c}^{T}}{\tau})\log{\frac{p(\frac{F_{i,c}^{T}}{\tau})}{p(\frac{F_{i,c}^{S}}{\tau})}}
\end{equation}

As defined by equation \eqref{eq2}, the teacher and student networks' outputs are denoted by $F^{S}$ and $F^{T}$, respectively. The relaxation parameter $\tau$ represents the temperature coefficient used to soften the outputs. For classification tasks, $W$ and $H$ take a value of 1, while for dense prediction, $W$ and $H$ refer to the output's width and height. $C$ always refers to the number of channels or classes of the outputs. The KL divergence between two logit outputs is represented by $\mathcal{KL}$.

\subsection{Balance Divergence Distillation}
To better leverage the softened ``dark knowledge'' from teacher, we propose a novel Balance Divergence Distillation (BDD) loss for Knowledge Distillation (KD). 
For a more comprehensive understanding, we analyze the calculation of KL divergence. By referring to equation \eqref{eq2} and the information theory in the paper\cite{shlens2014notes}, it can be easily deduced that the conventionally defined KL divergence is not a symmetric function:

\begin{equation}\label{eq3}
   \mathcal{KL}_f(p(F^T) || p(F^S)) \neq \mathcal{KL}_r(p(F^S) || p(F^T))
\end{equation}

In equation\eqref{eq3}, forward KL divergence is denoted by $\mathcal{KL}_f$, while reverse KL divergence is denoted by $\mathcal{KL}_r$. The forward KL divergence, also known as moment projection, is a commonly used method for KD \cite{hinton2015distilling}. The KD loss will be larger if the student network does not mimic the teacher's logit output well when calculating the forward KL divergence, i.e. if $p(F^T)>0$, the student network will be punished to encourage $p(F^S)>0$ \cite{murphy2012machine}.
However, the forward KL divergence typically emphasises the teacher's positive prediction where the probabilities are larger, while ignoring the calculation of the negative with extremely small probabilities.

\begin{equation}\label{eq4}
   \begin{aligned}
   \frac{\partial \mathcal{KL}_f(p(F^T) || p(F^S))}{\partial p(F^S)} &= 
   \partial \sum_{c=1}^C \bigg( p(\frac{F_{c}^{T}}{\tau}) \cdot \frac{\log{p(\frac{F_{c}^{T}}{\tau})}}{\log{p(\frac{F_{c}^{S}}{\tau})}}\bigg) / \partial p(F^S) \\
   &= \partial \sum_{c=1}^C \bigg(p(\frac{F_{c}^{T}}{\tau})\cdot \log{p(\frac{F_{c}^{T}}{\tau})} - p(\frac{F_{c}^{T}}{\tau})\cdot \log{p(\frac{F_{c}^{S}}{\tau})}\bigg) / \partial p(F^S) \\
   &= \partial \sum_{c=1}^C \bigg(-p(\frac{F_{c}^{T}}{\tau})\cdot \log{p(\frac{F_{c}^{S}}{\tau})}\bigg) / \partial p(F^S) \\
   \end{aligned}
\end{equation}

From the partial equation \eqref{eq4} for forward KL divergence, it becomes evident that the forward KL loss tends to disregard the influence of $p(F^S)$ as $p(F^T) \to 0$. In the training process of KD, avoiding a 0 value of $p(F^S)$ is important when using forward KL divergence. To achieve this, a small value, known as $\epsilon$ (e.g. $\epsilon=1e^{-12}$ in PyTorch), is commonly added. This small value $\epsilon$ restricts $p(F^S)$ from approaching 0, and may break the balance of knowledge transferred by $p(F^T)$, which means that if $p(F^T)$ is too small, as expressed in equation \eqref{eq5}, the loss may be ignored to mimic the logit output for the student.

\begin{equation}\label{eq5}
   \lim_{p(F_{i}^{T}) \to 0}\bigg(-p(F_{i}^{T}) \cdot \frac{\partial \log({p(\frac{F_{i}^{S}}{\tau})}+\epsilon)}{\partial p(\frac{F_{i}^{S}}{\tau})} \bigg)=0, \epsilon = 1e^{-12}
\end{equation}

\begin{figure}[htbp]
   \centering
   \begin{subfigure}[b]{0.45\textwidth}
      \centering
      \includegraphics[width=1.0\textwidth]{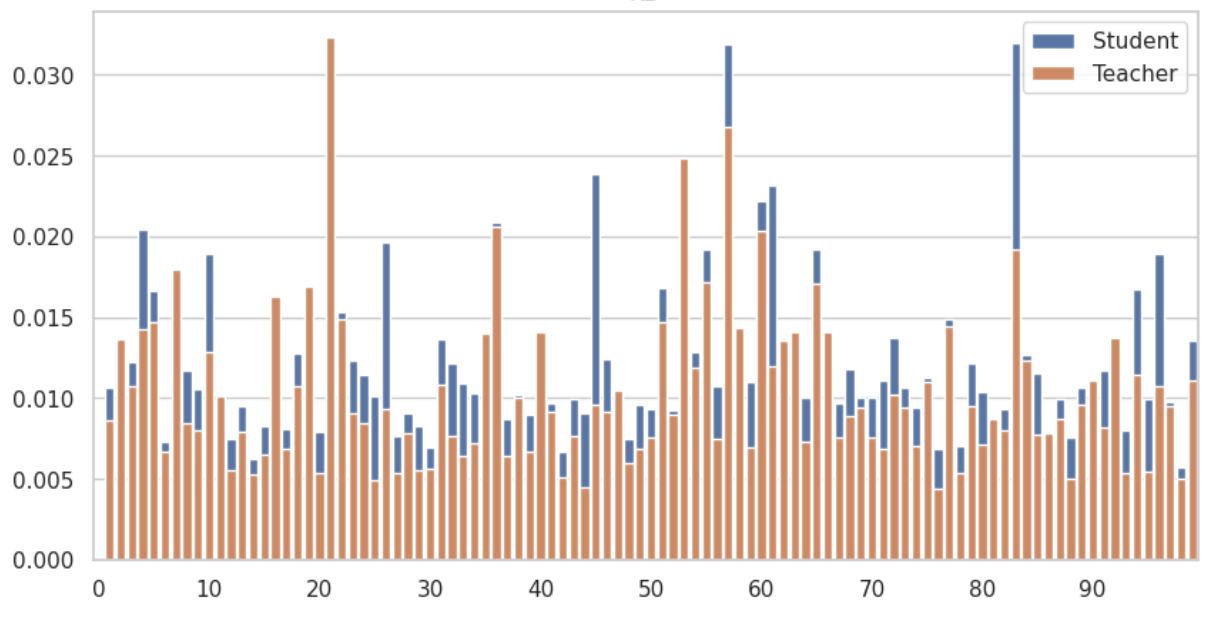}
      \caption{Output probability fitting by normal KD.}
      \label{fig:fig1a}
   \end{subfigure}
   \begin{subfigure}[b]{0.45\textwidth}
      \centering
      \includegraphics[width=1.0\textwidth]{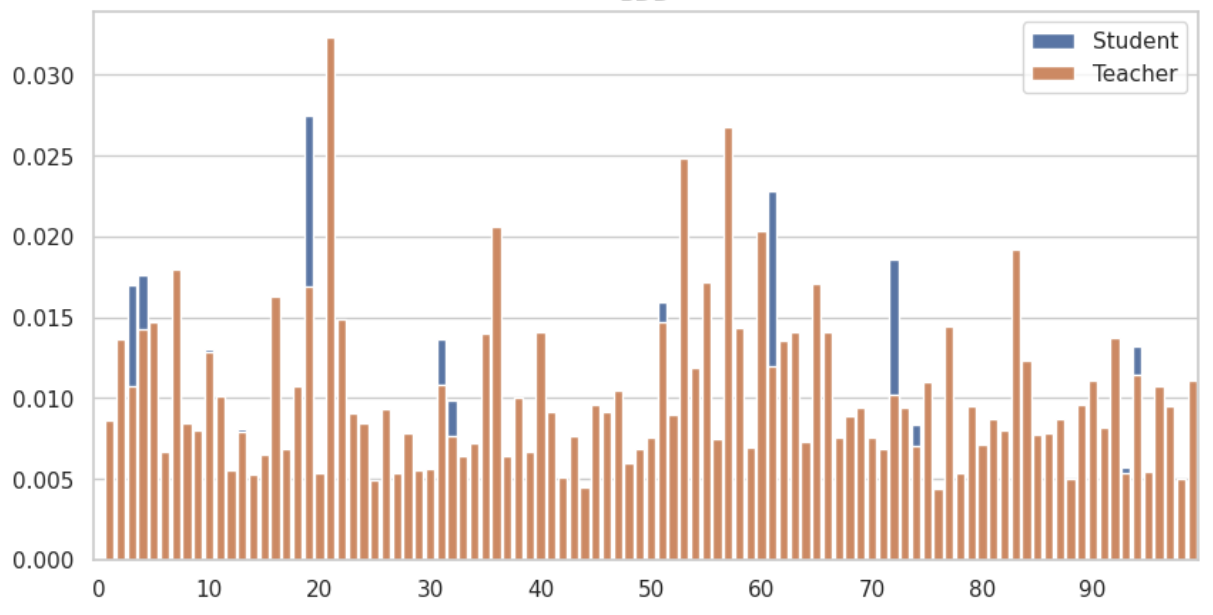}
      \caption{Output probability fitting by BDD.}
      \label{fig:fig1b}
   \end{subfigure}
   \caption{\textbf{Comparison of output probability fitting by normal KD and BDD}. This image illustrates the logit output fitting by KD and BDD on CIFAR100. The blue histogram shows the student's logit output, while the orange histogram represents the teacher's logit output. The left chart reveals that standard KD tends to overfit the positives of the teacher, neglecting the negatives and resulting in suboptimal learning outcomes. In contrast, the right figure demonstrates that BDD adeptly fits the teacher's negative regions. For enhanced visualization of minimal values, a temperature of 4.0 is applied for output smoothing and to clip the highest value.}
   \label{fig:fig1}
\end{figure}

This zero avoiding property of forward KL divergence means that the student network may overfit the positive samples of the teacher and ignore the negative samples, resulting in an imbalance of positive and negative sample learning. Especially a well pre-trained teacher network may predict accurately to distinguish between positives and negatives, often yielding extremely small values for negatives, which may lead to the student struggling to learn from the negative samples. From figure(\ref{fig:fig1}) we can see that the student network pays too much attention to the positive samples of the teacher, and ignores the learning of the negative samples, which leads to bad learning results and confirms our previous conclusion.

To tackle this imbalance problem, our Balance Divergence Distillation (BDD) loss employs reverse KL divergence, complementing the original forward KL divergence to address the imbalance between positive and negative samples. The term $\mathcal{KL}_r(p(F^S) || p(F^T))$ in equation \eqref{eq3}, known as reverse KL divergence or information projection, is a widely employed in variational inference\cite{ganguly2021introduction,neumann2011variational}. 
To ensure a balanced acquisition of knowledge from various forms of teacher supervision, we employ reverse KL divergence and implement the BDD loss, clearly defined as follows:

\begin{equation}
   \begin{aligned}
         \mathcal{L}_{BDD} &= \mathcal{KL}_f \Bigg({p(\frac{F^{T}}{\tau})}||{p(\frac{F^{S}}{\tau})}\Bigg) + \alpha \cdot \mathcal{KL}_r \Bigg({p(\frac{F^{S}}{\tau})}||{p(\frac{F^{T}}{\tau})} \Bigg) \\
         &=\frac{1}{C}\sum_{C}^{c=1} \Bigg(p(\frac{F_{c}^{T}}{\tau})\log{\frac{p(\frac{F_{c}^{T}}{\tau})}{p(\frac{F_{c}^{S}}{\tau})}} + \alpha \cdot p(\frac{F_{c}^{S}}{\tau})\log{\frac{p(\frac{F_{c}^{S}}{\tau})}{p(\frac{F_{c}^{T}}{\tau})}} \Bigg)
   \end{aligned}
\end{equation}

The proposed BDD loss can be acknowledged as a balanced loss that incorporates both positive and negative samples by including the $\mathcal{KL}_r$ formula, since the forward part may focuses on large probabilities while the reverse part may focuses on extremely small values. In equation \eqref{eq7}, it is evident that when the probability $p(F^T)$ becomes exceedingly small, the probability of the student $p(F^S)$ remains unaffected in reverse KL divergence. Consequently, BDD addresses the problem of imbalance between positive and negative samples observed in forward KL divergence.

\begin{equation}\label{eq7}
   \begin{aligned}
      \partial \mathcal{L}_{BDD} / \partial p(F^S) &= \partial \Bigg(\mathcal{KL}_f \bigg({p(\frac{F^{T}}{\tau})}||{p(\frac{F^{S}}{\tau})}\bigg) + \alpha \cdot \mathcal{KL}_r \bigg({p(\frac{F^{S}}{\tau})}||{p(\frac{F^{T}}{\tau})} \bigg)\Bigg) /\partial p(F^S) \\
      & = \partial \sum_{c=1}^C \bigg(-p(\frac{F_{c}^{T}}{\tau})\cdot \log{p(\frac{F_{c}^{S}}{\tau})}\bigg) / \partial p(F^S) + \\
      & \alpha \cdot \partial \sum_{c=1}^C \Bigg(({p(\frac{F_{c}^{S}}{\tau}) \cdot \log p(\frac{F_{c}^{S}}{\tau}) } ) - ({p(\frac{F_{c}^{S}}{\tau}) \cdot \log p(\frac{F_{c}^{T}}{\tau}) } ) \Bigg)  /\partial p(F^S) \\
   \end{aligned}
\end{equation}

The parameter $\alpha$ serves as a dynamic hyperparameter, facilitating the balance between forward and reverse KL divergence. The integration of the BDD loss into the existing KD framework is seamless and straightforward.

\subsection{Balance of Temperature Between Forward and Reverse}
To better leverage the balance in BDD, we investigated its effectiveness on both traditional KD and state-of-the-art (SOTA) logit-based distillation methods in classification \cite{zhao2022decoupled,Jin_2023_CVPR,li2023curriculum}. We observed that a simple summation of forward and reverse KL divergence did not yield optimal outcomes.

In the reverse calculation section of equation \eqref{eq7}, it is noticeable that even though reverse KL divergence encourages the sampling of negative samples as $p(F^T) \to 0$, the calculated value of positive samples may also increase when both teacher and student outputs are high.

From previous works, it is evident that with the guidance of label information or the augmentation of predictions, further accentuations are achieved by balancing the effects of positive and negative samples.  These methods with high computational costs play a crucial role in distinguishing the KD effects among different target samples during the training process. 
Inspired by temperature effects, if additional supervision information is not desired, a balance between positive and negative samples can be achieved by adjusting the temperature coefficient.
thus we enhance the BDD loss using different temperatures as a form of augmentation balance. Equation \eqref{eq8} defines the improved BDD loss:

\begin{equation}\label{eq8}
   \begin{split}
     \mathcal{L}_{BDD} &= \frac{1}{C}\sum_{c=1}^{C}\Bigg(\mathcal{KL}_f({p(\frac{F_{c}^{T}}{\tau_f})}||{p(\frac{F_{c}^{S}}{\tau_f})}) + \alpha \cdot \mathcal{KL}_r({p(\frac{F_{c}^{S}}{\tau_r})}||{p(\frac{F_{c}^{T}}{\tau_r})})\Bigg)
   \end{split}
 \end{equation}

The $\tau_f$ denotes the temperature coefficient applied to the forward KL divergence, while $\tau_r$ is associated with the reverse KL divergence.
In general, higher temperatures in the softmax produce a more evenly distributed output, reducing the certainty of the distribution while lower temperatures have the opposite effect. 
When the temperature for forward KL divergence is high and that for reverse KL divergence is low, it further weakens the penalty on positive samples and strengthens the penalty on negative samples. 
However, if the temperature for forward KL divergence is low and the temperature for reverse KL divergence is high, it reduces the penalty on negative samples and amplifies the penalty on positive samples. 
By using different temperature coefficients to regulate the enhancement and reduction of positive and negative aspects in BDD, a more refined balance is achieved, eliminating the need for additional labels as references. 

To simplify the calculation, the temperature is accumulated as an integral term in the divergence calculation, which takes advantage of the temperature's ability to balance the positive and negative samples. The refined BDD loss is defined as follows:

\begin{equation}\label{eq9}
   \begin{aligned}
      \mathcal{L}_{BDD} &= \frac{1}{C}\sum_{c=1}^{C} \Bigg(\int_{\tau_f} p(\frac{F_{c}^{T}}{\tau_f})\log\frac{p(\frac{F_{c}^{T}}{\tau_f})}{p(\frac{F_{c}^{S}}{\tau_f})}d\frac{F_{c}^{T}}{\tau_f} 
                        + \alpha \cdot \int_{\tau_r}p(\frac{F_{c}^{S}}{\tau_r})\log\frac{p(\frac{F_{c}^{S}}{\tau_r})}{p(\frac{F_{c}^{T}}{\tau_r})}d\frac{F_{c}^{S}}{\tau_r}\Bigg)
   \end{aligned}
\end{equation}


\subsection{Effects of BDD on Dense Prediction} 
To further demonstrate the effectiveness of BDD not only in classification tasks but also in dense prediction tasks, we applied BDD to semantic segmentation tasks. 
In the area of semantic segmentation, previous work such as CWD \cite{shu2021channel} has shown that mimicking channel-wise attention statistics by normalizing the output of each channel is more effective than mimicking the logit output pixel by pixel. 
However, the variation of the distributional characteristics within the logit output layer presents a challenge for the KL divergence distillation method used previously, making it more difficult to accurately capture the distributional state of the teacher network.

\begin{figure}[htbp]
   \centering
   \begin{subfigure}[b]{0.45\textwidth}
      \centering
      \includegraphics[width=1\textwidth]{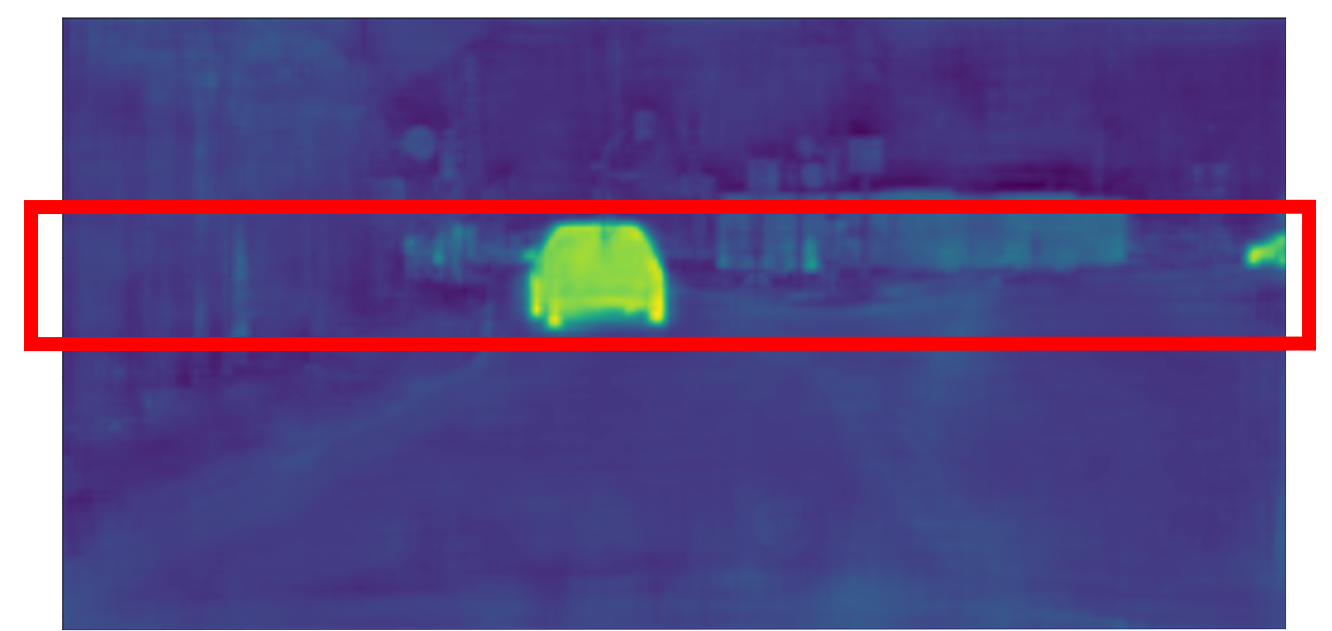}
      \caption{Output feature map of semantic segmentation.}
      \label{fig:fig2a}
   \end{subfigure}
   \begin{subfigure}[b]{0.45\textwidth}
      \centering
      \includegraphics[width=1\textwidth]{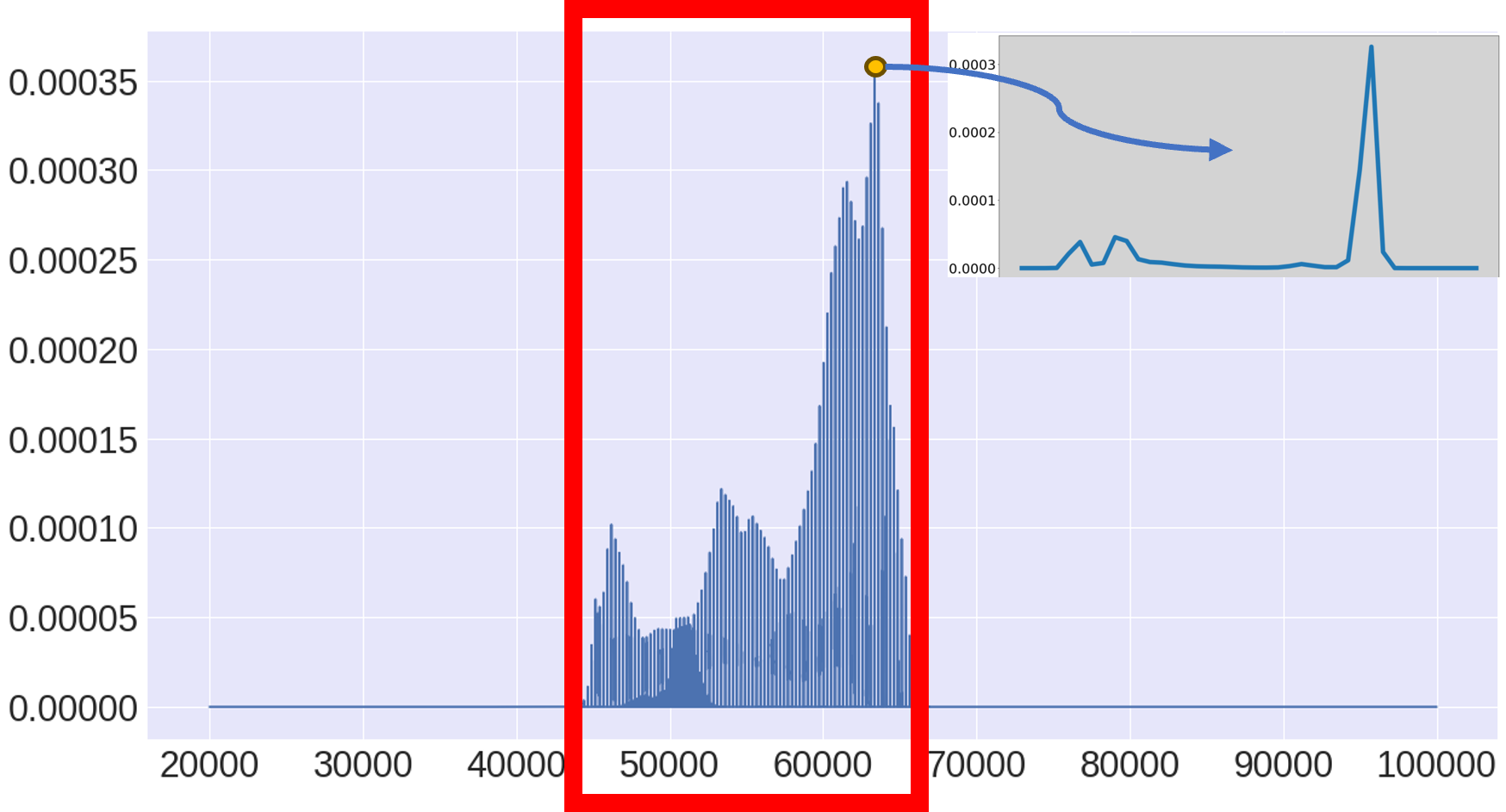}
      \caption{Channel-wise attention output probability.}
      \label{fig:fig2b}
   \end{subfigure}
   \caption{\textbf{Unbalance in the output feature map of semantic segmentation}. The image shows output feature maps and channel attention probability maps for a semantic segmentation task on the Cityscapes dataset. The red boxes at the top and bottom of the image represent the selected regions' output features and their corresponding channel attention probability distributions. The vertical axis of the channel attention indicates the probability values, while the horizontal axis represents the flattened pixel values.}
   \label{fig:fig2}
\end{figure}

The data presented in figure(\ref{fig:fig2}) shows that the logit output layer of CWD exhibits a visible imbalance between positive and negative samples. For example, when predicting vehicle positions, the corresponding prediction channel shows a greater emphasis on the exact location of the vehicles, resulting in a more significant imbalance.
Therefore, we applied BDD to the channel-wise attention output in semantic segmentation tasks and defined its loss as:

\begin{equation}\label{eq10}
   \begin{split}
      \mathcal{L}_{BDD_{seg}} &= \frac{1}{C}\sum_{c=1}^{C}\sum_{i=1}^{H \times W} \Bigg(p(\frac{F_{c,i}^{T}}{\tau_f})\log{\frac{p(\frac{F_{c,i}^{T}}{\tau_f})}{p(\frac{F_{c,i}^{S}}{\tau_f})}} + \alpha \cdot p(\frac{F_{c,i}^{S}}{\tau_r})\log{\frac{p(\frac{F_{c,i}^{S}}{\tau_r})}{p(\frac{F_{c,i}^{T}}{\tau_r})}}\Bigg)
   \end{split}
\end{equation}

Through the computation of equation \eqref{eq10} and comparative experiments, the BDD was found to be more balanced for channel-wise attention logit output than normal KL loss, and can achieve certain performance improvements for different types of attention mechanisms in semantic segmentation tasks, proving the effectiveness and robustness of BDD.

\subsection{Overall loss}
In summary, an attempt is made to add the corresponding BDD loss to various knowledge distillation methods. The final loss definition is achieved by combining the BDD loss with other losses, leading to the following result:
   \begin{equation}
      \mathcal{L}_{overall} = \mathcal{L}_{CE} + \beta \cdot (\underbrace{\mathcal{KL}_{f} + \alpha \cdot \mathcal{KL}_{r}}_{\mathcal{L}_{BDD}})
   \end{equation}

Where $ \mathcal{L}_{CE}$ is the cross-entropy loss, and the sum of the forward KL divergence and the reverse KL divergence is the BDD loss, $\mathcal{L}{overall}$ is the final loss. The BDD loss is only calculated on the final attention output, making it easy to integrate into different tasks. Algorithm\ref{alg:alg1} shows the pseudo-code of training process with BDD.

\begin{algorithm}
   \caption{Pseudo-code of BDD in pytorch style}
   \label{alg:alg1}
   \begin{algorithmic}
      \Procedure{BDD}{$feat_{student}, feat_{teacher}, T_f, T_r, \alpha$}
      \State $p_{student_f} = \text{F.softmax}(feat_{student} / T_f)$
      \State $p_{student_r} = \text{F.softmax}(feat_{student} / T_r)$
      \State $p_{teacher_f} = \text{F.softmax}(feat_{teacher} / T_f)$
      \State $p_{teacher_r} = \text{F.softmax}(feat_{teacher} / T_r)$
      \State $\mathcal{KL}_f = \text{F.kl\_div}(\log p_{teacher_f} , p_{student_f})$
      \State $\mathcal{KL}_r = \text{F.kl\_div}(\log p_{student_r} , p_{teacher_r})$
      \State $\mathcal{L}_{BDD} \gets \mathcal{KL}_f + \alpha \cdot \mathcal{KL}_r$
      \State \Return $\mathcal{L}_{BDD}$
      \EndProcedure
   \end{algorithmic}
\end{algorithm}

\section{Experiments}
In this section, we provide a detailed description of the datasets used in different tasks, experimental settings, and analysis of experimental results. We also compare our method with other state-of-the-art classification and semantic segmentation methods, demonstrating the effectiveness of our method in both different tasks of classification and dense predictions.
\subsection{Experimental Setup}
\textbf{Datasets}. We used different datasets for classification and semantic segmentation tasks.

CIFAR-100 \cite{krizhevsky2009learning} is a dataset for image classification. It contains 100 classes with 600 images per class, where 500 images are used for training and 100 images are used for testing. We use $32\times32$ as input resolution , the learning rate is 0.025 or 0.05, and we trained for 240 epochs with a batch size of 64.

ImageNet \cite{russakovsky2015imagenet} is a large-scale dataset for image classification. It contains 1000 classes with 1.2 million images for training and 50,000 images for validation. The input resolution for training is $224\times224$, the learning rate is 0.1, and we trained for 100 epochs with a batch size of 128.

Cityscapes \cite{cordts2016cityscapes} dataset is employed for the purpose of semantic urban scene understanding. It comprises 30 common classes and a total of 5000 images, each image has a size of $2048\times1024$ pixels and is sourced from 50 different cities. During training, 19 classes are actively utilized. Specifically, there are 2975 images for training, 500 for validation, and 1525 for testing, the coarsely labeled data is excluded from our experiments. The input resolution for the images is set to $512\times512$ pixels for training, while the original resolution is maintained for validation and testing. The learning rate used is 0.01, and the training process spans 40000 iterations with a batch size of 8.

\textbf{Evaluation metrics}. To assess the effectiveness and efficiency of our proposed BDD method for both classification and segmentation tasks, we follow the methodology established in previous research \cite{zhao2022decoupled,shu2021channel,liu2019structured}. For the classification task, we use the top-1 accuracy as the evaluation metric. For the semantic segmentation task, we use the mean intersection over union (mIoU) as the evaluation metric. 

\textbf{Implementation details}.
For the classification Knowledge Distillation (KD) method, we utilize different networks for the teacher and student networks. The teacher network primarily consists of ResNet \cite{he2016identity} architectures (such as ResNet32$\times$4 and ResNet56), WideResNet \cite{DBLP:conf/bmvc/ZagoruykoK16} (WRN) structures (e.g., WRN-40-2), and VGG \cite{DBLP:journals/corr/SimonyanZ14a} architectures (like VGG13). In contrast, the student network is designed to be a compressed version corresponding to the teacher network, which comprises models like resnet20, resnet8$\times$4, WRN-16-2, WRN-40-1, VGG8, and the lightweight MobileNetV2 \cite{sandler2018mobilenetv2} ShuffleNetV1 \cite{zhang2018shufflenet} ShuffleNetV2 \cite{ma2018shufflenet}.

For the semantic segmentation KD method, we use PSPNet-R101 \cite{zhao2017pyramid} as the teacher network and train student networks with the PSPNet and Deeplab \cite{chen2017deeplab} backbones based on ResNet18. We conduct ablation experiments to compare the effects of temperature coefficients in different tasks. The hyperparameter alpha is generally set to 4.0. For typical network architectures, the value of the temperature coefficient in the forward KL divergence is 2.0, whereas in the reverse KL divergence it is 8.0. Furthermore, in situations where there is greater emphasis on knowledge distillation loss in semantic segmentation tasks, we implement the method described in the paper by CWD \cite{shu2021channel}, which involves setting the scaling factor to 3.0.

\subsection{Experiments on Classification}

Firstly, we compared the efficacy of diverse distillation methods in classification assignments, encompassing logit-based distillation procedures like KD \cite{hinton2015distilling}, DKD (Decoupled Knowledge Distillation) \cite{zhao2022decoupled} and MLD (Multi-level Logit Distillation) \cite{Jin_2023_CVPR}. Additionally, feature-based distillation methods, such as FitNet \cite{DBLP:journals/corr/RomeroBKCGB14}, RKD (Relational Knowledge Distillation) \cite{park2019relational}, AT (Attention Transfer) \cite{DBLP:conf/iclr/ZagoruykoK17}, CRD (Contrastive Representation Distillation) \cite{DBLP:conf/iclr/TianKI20}, OFD (Overhaul Feature Distillation) \cite{heo2019comprehensive}, and ReviewKD \cite{chen2021distilling}, were also utilized in all experiments. 

\begin{table*}[htbp]
   \centering
   \resizebox{1.0\textwidth}{!}{
      \begin{tabular}{cc|cccccc}
         \hline
         \multirow{4}{*}{methods} & \multirow{2}{*}{teacher} &  ResNet56 & ResNet101 & ResNet32$\times$4 & WRN-40-2& WRN-40-2& VGG13 \\
                                  &                          &  72.34    &  74.31    &  79.42            &  75.61  &  75.61  &  74.64   \\
                                  & \multirow{2}{*}{student} &  ResNet20 & ResNet32  & ResNet8$\times$4  & WRN-16-2& WRN-40-1& VGG8  \\
                                  &                          &  69.06    &  71.14    &  72.50            &  73.26  &  71.98  &  70.36    \\ \hline
         & FitNet                                            &  69.21    &  71.06    &  73.50            &  73.58  &  72.24  &  71.02    \\
         & RKD                                               &  69.61    &  71.82    &  71.90            &  73.35  &  72.22  &  71.48    \\
         features & CRD                                      &  71.16    &  73.48    &  75.51            &  75.48  &  74.14  &  73.94    \\
         & OFD                                               &  70.98    &  73.23    &  74.95            &  75.24  &  74.33  &  73.95    \\
         & ReviewKD                                          &  71.89    &  73.89    &  75.63            &  76.12  &  75.09  &  74.84    \\ \hline
      \multirow{4}{*}{logit} & DKD                           &  71.97    &  74.11    &  76.32            &  76.24  &  74.81  &  74.68    \\
         & KD(baseline)                                      &  70.66    &  73.08    &  73.33            &  74.92  &  73.54  &  72.98    \\
         & Ours                                              &  72.00    &  74.17    &  76.21            &  75.72  &  74.68  &  74.74    \\
         & $\Delta$                                          &  +1.34    &  +1.09    &  +1.29            &  +0.8   &  +1.14  &  +1.76    \\ \hline
      \multirow{2}{*}{multi-level logit} & MLD               &  72.19    &  74.11    &  77.08            &  76.63  &  75.35  &  75.18    \\
         & Ours$^*$                                          &  \textbf{72.35}  &  \textbf{74.29}  &  \textbf{77.22}  &  \textbf{76.85}  &  \textbf{75.56}  &  \textbf{75.25}   \\ \hline
      \end{tabular}
   }
   \caption{\textbf{Top-1 accuracy(\%) on CIFAR100 validation}. This table is a comparison of different distillation methods with homogeneous network architectures between teacher and student. The symbol $\Delta$ indicates the performance improvement over the baseline, * indicates that the BDD is trained based on MLD. All results are the average of five runs.}
   \label{tab:samearch}
\end{table*}

\begin{table}
   \centering
   \resizebox{1.0\textwidth}{!}{
   \begin{tabular}{cc|ccccc}
      \hline
      \multirow{4}{*}{methods}& \multirow{2}{*}{teacher}  & ResNet32$\times$4 & WRN-40-2 &  VGG13         & ResNet50    & ResNet32$\times$4 \\
                              &                           &        79.42      & 75.61    &  74.64         &  79.34      & 79.42             \\
                              & \multirow{2}{*}{student}  &  ShuffleNetV1     &ShuffleNetV1& MobileNetV2  & MobileNetV2 & ShuffleNetV2      \\
                              &                           &        70.50      & 70.50    &  64.60         &  64.60      & 71.82             \\ \hline
      \multirow{5}{*}{feature}&          FitNet           &        73.59      & 73.73    &  64.14         &  63.16      & 73.54             \\
                              &          RKD              &        72.28      & 72.21    &  64.52         &  64.43      & 73.21             \\
                              &          CRD              &        75.11      & 76.05    &  69.73         &  69.11      & 75.65             \\
                              &          OFD              &        75.98      & 75.85    &  69.48         &  69.04      & 76.82             \\
                              &          ReviewKD         & \textbf{77.45}    & 77.14    &  70.37         &  69.89      & 77.78             \\ \hline
      \multirow{4}{*}{logit}  &          DKD              &        76.45      & 76.70    &  69.71         &  70.35      & 77.07             \\
                              &          KD(baseline)     &        74.07      & 74.83    &  67.37         &  67.35      & 74.45             \\ 
                              &          Ours             &        76.01      & 76.54    &  69.52         &  70.40      & 76.88             \\
                              &          $\Delta$         &        +1.94      & +1.71    &  +2.15         &  +3.05      & +2.43             \\ \hline
\multirow{2}{*}{multi-level logit} &     MLD              &        77.18      & 77.44    &  70.57         &  71.04      & 78.44            \\
                              &          Ours$^*$         &        77.24 & \textbf{77.61}&\textbf{70.71}  &\textbf{71.25}& \textbf{78.53}   \\ \hline
   \end{tabular}}
   \caption{\textbf{Top-1 accuracy(\%) on CIFAR100 validation}. This table is a comparison of different distillation methods with a heterogeneous network between teacher and student. The symbol $\Delta$ indicates the performance improvement over the baseline, * indicates that the BDD is trained combined with MLD. All results are the average of five runs.}
   \label{tab:differentarch}
\end{table}

In all of the experiments, traditional cross-entropy loss was employed in conjunction with distillation loss to improve the performance of the student network. Table \ref{tab:samearch} presents a comparison of various distillation methods using identical teacher and student network architectures. Our results indicate that BDD demonstrated consistent enhancements in all teacher-student pairs compared to the traditional KD methods with about $1\% \sim 2\%$ top-1 accuracy improvements. Additionally, BDD even showed some improvements when compared to DKD, the current state-of-the-art logit-based distillation technique.

To better verify the effectiveness of BDD, we compared the performance of distillation methods when the teacher and student networks had varying architectures. As table \ref{tab:differentarch} shown, our method consistently achieved a performance improvement of around $2\% \sim 3\%$ in this comparison, providing evidence for the effectiveness of our approach. 

\begin{table}
   \centering
   \resizebox{0.7\textwidth}{!}{
   \begin{tabular}{cc|ccccc}
      \hline
      \multirow{4}{*}{methods}    &  \multirow{2}{*}{teacher}  &   &  ResNet34      & ResNet50       \\
                                  &                            &   &  73.31         &  76.16         \\
                                  &  \multirow{2}{*}{student}  &   &  ResNet18      & MobileNetV2    \\
                                  &                            &   &  69.75         &  68.87         \\ \hline
      \multirow{4}{*}{feature}    &   AT                       &   &  70.69         &  69.56         \\
                                  &   CRD                      &   &  71.17         &  71.37         \\
                                  &   OFD                      &   &  70.81         &  71.25         \\
                                  &   ReviewKD                 &   &  71.61         &  72.56         \\ \hline
      \multirow{4}{*}{logit}      &   DKD                      &   &  71.70         &  72.05         \\
                                  &   KD(baseline)             &   &  70.66         &  68.58         \\ 
                                  &   Ours                     &   &  71.55         &  71.83         \\ \hline
\multirow{2}{*}{multi-level logit}&   MLD                      &   &  71.90         &  73.01         \\
                                  &   Ours$^*$                 &   & \textbf{71.97} & \textbf{73.12} \\ \hline
   \end{tabular}}
   \caption{\textbf{Top-1 accuracy(\%) on ImageNet}. We use ResNet34 and ResNet18 as backbone for homogeneous architecture knowledge distillation, and use ResNet50 and MobileNetV2 as backbone for heterogeneous architecture knowledge distillation. * indicates that the BDD is trained as part of MLD}
   \label{tab:imagenet}
\end{table}

Comparing the multi-level logit method in table \ref{tab:samearch} and table \ref{tab:differentarch}, when we add BDD to the MLD method and keep the training environments consistent (increasing the data augmentation and training epochs), BDD may achieve better results, showing that our method can be easily combined with other distillation methods. The results of ImageNet in the table \ref{tab:imagenet} also show that even on large datasets, our method can still achieve better results, competing with other distillation methods.

\subsection{Experiments on Dense Prediction}
As the methods discussed in previous works, we also conducted experiments on semantic segmentation compared with other dense prediction knowledge distillation methods, channel-wise distillation CWD \cite{shu2021channel}, relation-based knowledge distillation method IFVD \cite{wang2020intra} and feature-based knowledge distillation method MGD \cite{yang2022masked}. 
We used the Cityscapes dataset to evaluate the performance of our method. The input resolution for the images is set to 512$\times$512 pixels for training, while the original resolution is maintained for validation. The teacher network is PSPNet-R101, and the student network are PSPNet-R18 and DeepLabV3-R18. The results are shown in table \ref{tab:segmentation}. We can see that our method can also achieve a certain improvement compared with the baseline student network PSPNet-R18 (4.72 mIoU) and DeepLabV3-R18 (2.73 mIoU). By adding the BDD loss on channel-wise attention, only use logit outputs can bring about 0.58 mIoU improvement compared with CWD. Compared with other knowledge distillation methods, our method achieve the best performance on both same architecture and different architecture. 

\begin{table}[htbp]
   \begin{center}
      \resizebox{0.7\textwidth}{!}{
         \begin{tabular}{l|c|c|c|c}
            \hline
            Method               & Input size   & Params(M) & FLOPs(G) & mIoU            \\ \hline
            PSPNet-R101(T)       &512$\times$512& 70.43     & 574.9    & 78.50           \\ 
            PSPNet-R18(S)        &512$\times$512& 13.07     & 125.8    & 70.90           \\ \hline
            IFVD                 &512$\times$512& 13.07     & 125.8    & 74.54           \\
            MGD                  &512$\times$512& 13.07     & 125.8    & 73.63           \\
            CWD-logit            &512$\times$512& 13.07     & 125.8    & 74.87           \\
            Ours                  &512$\times$512& 13.07    & 125.8    & \textbf{75.62}  \\
            $\Delta$             & -            & -         &  -       & +4.72           \\ \hline
            PSPNet-R101(T)       &512$\times$512& 70.43     & 574.9    & 78.50           \\ 
            DeepLabV3-R18(S)     &512$\times$512& 12.62     & 123.9    & 73.37           \\ \hline
            IFVD                 &512$\times$512& 12.62     & 123.9    & 74.09           \\
            MGD                  &512$\times$512& 12.62     & 123.9    & 76.02           \\
            CWD-logit            &512$\times$512& 12.62     & 123.9    & 75.91           \\
            Ours                  &512$\times$512& 12.62     & 123.9    & \textbf{76.31}  \\  
            $\Delta$             & -            &  -        &  -       & +2.94           \\ \hline
         \end{tabular}
      }
      \caption{ \textbf{Comparison of semantic segmentation knowledge distillation}. We utilised PSPNet-R101 as a teacher, PSPNet-R18 and DeepLabV3-R18 as students with different segmentation methods for knowledge distillation. The obtained results represent the average of three runs.}
      \label{tab:segmentation}
   \end{center}
\end{table}

\begin{figure}[htbp]
   \centering
   \begin{subfigure}[t]{1.0\textwidth}
      \includegraphics[width=\textwidth]{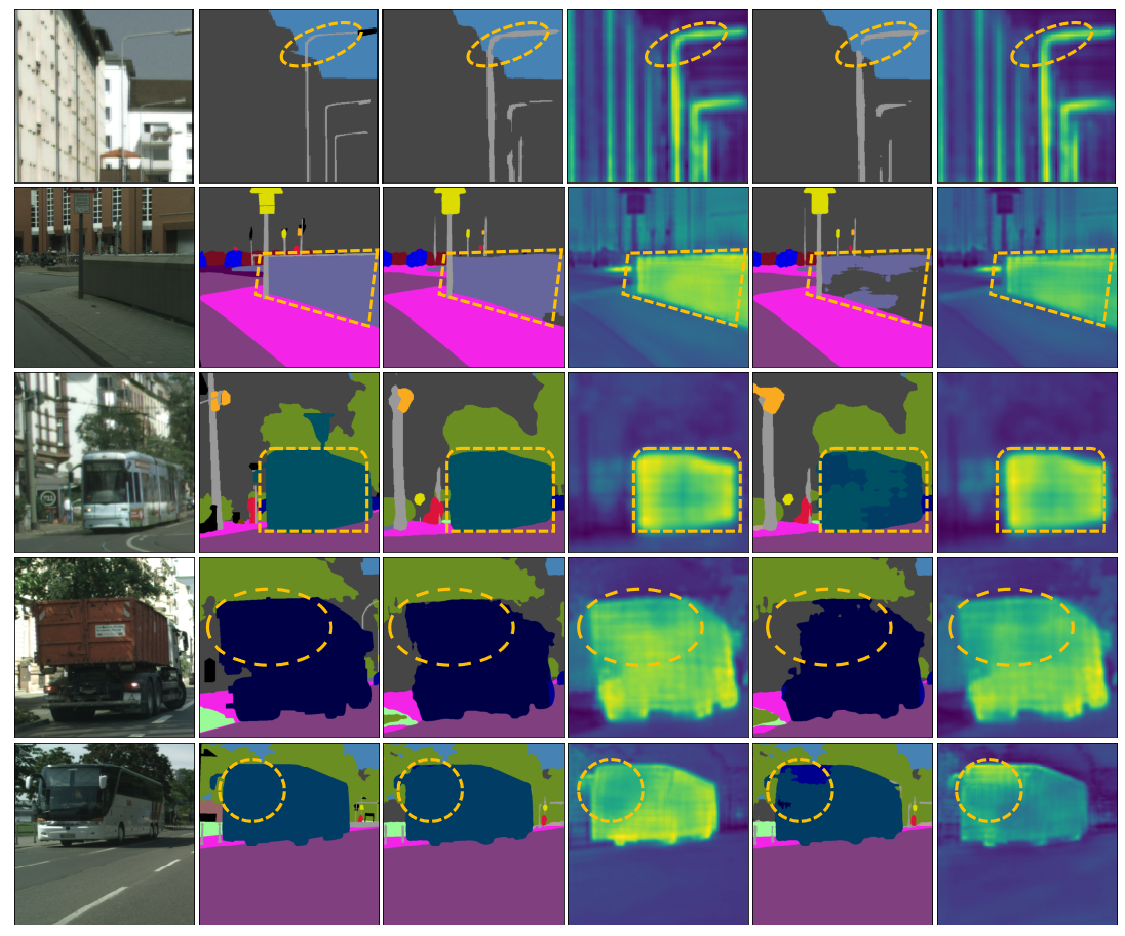}
      \scriptsize
      \begin{tabular}{@{}cccccc@{}}
         \phantomsubcaption \qquad\quad(a) Image & \phantomsubcaption \qquad\quad(b) GT &  \phantomsubcaption \qquad\quad(c) BDD & \phantomsubcaption \qquad\quad(d) BDD$^*$ & \phantomsubcaption \qquad\quad(e) CWD & \phantomsubcaption \qquad\quad(f) CWD$^*$
      \end{tabular}
   \end{subfigure}
   \caption{\textbf{Qualitative segmentation results and channel distributions}. This figure displays the output feature maps and channel attention probability maps of PSPNet-R18 for a semantic segmentation task on the Cityscapes dataset. (a) raw images, (b) ground truth(GT), (c)our method(BDD), (d)channel attention probability of our method(BDD$^*$), (e)channel-wise distillation(CWD), (f)channel attention probability of CWD(CWD$^*$). The selected regions indicate the segmentation quality of the different knowledge distillation methods.}
   \label{fig:fig4}
\end{figure}

In addition, we report the full class IoU of both our method and two recent knowledge distillation methods for sementatic segmentation in Table \ref{tab:classiou}. Our method significantly improves the class accuracy of several objects, such as sidewalk, traffic light, truck, bus and train, indicating that our BDD method may transfer the dense knowledge well. From figure(\ref{fig:fig4}), it is evident that our method outperforms other methods in terms of segmentation results. Additionally, the channel attention probability maps of our method display a closer similarity to the ground truth.

\begin{table}
   \centering
   \resizebox{1.0\textwidth}{!}{
   \begin{tabular}{c|c|c|c|c|c|c|c|c|c|c}
      \hline
      Mehtod &  mIoU          &  road         &   sidewalk    &   building    &    wall      &    fence      &     pole     &  traffic light &  traffic sign &   vegetation   \\ \hline
       IFVD  &  71.66         & 97.56         &   81.44       &   91.49       &   44.45      &   55.95       &    62.40     &  66.38         &  76.44        &   91.85        \\
       CWD   &  74.72         & 97.56         &   81.61       &   91.82       &   47.28      &   57.05       &\textbf{62.50}&  67.77         &  77.02        &\textbf{92.10}  \\
       Ours  & \textbf{75.62} &\textbf{97.68} &\textbf{82.33} &\textbf{91.91} &\textbf{48.17}&\textbf{57.07} &    62.29     &\textbf{68.47}  &\textbf{77.46} &   92.08        \\ \hline
       Class &terrain         & sky           &   person      &   rider       &   car        &   truck       &    bus       &  train         &  motorcycle   &   bicycle      \\ \hline
       IFVD  &  61.29         & 93.97         &   78.64       &   52.33       &   93.50      &   60.25       &    74.70     &  58.81         &  44.85        &   75.41        \\
       CWD   &  63.4          &\textbf{94.32} &\textbf{80.04} &\textbf{58.82} &   94.17      &   68.7        &    85.04     &  71.63         &\textbf{52.34} &\textbf{76.52}  \\
       Ours  &\textbf{64.35}  & 94.29         &   79.83       &   58.67       &\textbf{94.32}&\textbf{74.18} &\textbf{86.94}&\textbf{78.60}  &  52.00        &   76.22        \\ \hline
   \end{tabular}}
   \caption{\textbf{Class IoU of Cityscapes}. We compared our method with other two typical knowledge distillation methods on the validation set of Cityscape for each class IoU, PSPNet-R101 and PSPNet-R18 were used as teacher and student network respectively.}
   \label{tab:classiou}
\end{table}

\subsection{Ablation Study}

In this section we conduct ablation experiments to verify the effectiveness of BDD. We investigate the effects of the hyperparameters with different temperature coefficients in training procedures. The baseline student networks are ResNet8$\times$4 and VGG8, the teacher networks are ResNet32$\times$4 and VGG13. 

\textbf{Effectiveness of forward and reverse KL-divergence.} Our inquiry commences by evaluating the efficacy of the forward and reverse KL-divergence in the BDD framework. We execute a comparative analysis of BDD's performance with and without reverse KL-divergence, aided by regulating the hyperparameter $\alpha$. The findings, as displayed in table \ref{tab:ablationalpha}, demonstrate a notable pattern: the efficiency of knowledge distillation is limited when reverse KL-divergence is not integrated, in comparison to BDD with such a feature. It is important to acknowledge, however, that an exceedingly high value of $\alpha$ may result in a reduction of accuracy. This highlights the significance of striking a suitable balance in employing reverse KL-divergence. An optimal scaling of this divergence term can considerably improve the overall performance of BDD.

\textbf{Impact of the temperature.} 
Temperature coefficient plays a crucial role in balancing the training process between forward and reverse KL-divergences. Through our ablation experiments shown in table \ref{tab:ablationtemperature}, we have observed significant variations in the temperature ratios associated with models of different architectures. As evident from the results presented in table \ref{tab:Classification}, this temperature-controlled approach leads to superior performance, surpassing the current state of the art methods in logit knowledge distillation. 

\begin{table}[htbp]
   \centering
   \resizebox{0.5\textwidth}{!}{
      \begin{tabular}{c|c|c|c|c|c}
         \hline
         student & KD & BDD & BDD$^+$ & top-1 & $\Delta$ \\
         \hline
         \multicolumn{5}{c}{ResNet32$\times$4 as teacher} \\
       \hline
       \multirow{4}{*}{ResNet8$\times$4} 
       &      -     &     -      &      -     & 72.50 & -     \\
       & \checkmark &            &            & 73.15 & +0.65 \\
       & \checkmark & \checkmark &            & 75.71 & +3.21 \\
       & \checkmark &            & \checkmark & 76.21 & +3.71 \\
       \hline
       \multicolumn{5}{c}{VGG13 as teacher} \\
       \hline
       \multirow{4}{*}{VGG8} 
       &    -       &   -        &     -      & 70.36 & -     \\
       & \checkmark &            &            & 72.98 & +2.62 \\
       & \checkmark & \checkmark &            & 74.23 & +3.87 \\
       & \checkmark &            & \checkmark & 74.74 & +4.38 \\
       \hline
   \end{tabular}}
   \caption{\textbf{Classification Experiments on CIFAR-100}. top-1 represents the accuracy (\%) of student networks on validation set. The symbol $\Delta$ denotes the improvement in performance compared to the baseline. BDD$^+$ represents the BDD loss with different temperature coefficients.}
   \label{tab:Classification}
\end{table}

For instance, in models with larger capacity and easier learning, such as the ResNet structure, the teacher network achieves higher accuracy. This requires a softer supervision during the forward process and a stronger emphasis on learning from minima during the reverse process. Consequently, we set $\tau_f$ to a smaller value and $\tau_r$ to a larger value. Conversely, in simpler model structures like VGG, where the teacher's accuracy is relatively lower from the outset, it becomes imperative to capture more explicit information from positive samples while reducing the emphasis on minima. Hence, we opt for a larger $\tau_f$ and a smaller $\tau_r$.

\begin{table}
   \centering
   \resizebox{0.5\textwidth}{!}{
   \begin{tabular}{c|c|c|c|c|c}
      \hline
      \multirow{2}{*}{$\alpha$} & \multicolumn{5}{c}{ResNet32$\times$4 $\to$ ResNet8$\times$4} \\
      \cline{2-6}
      &  $0.0$  &  $1.0$  &  $2.0$  &  $4.0$ & $8.0$  \\ \hline
      top-1             &  73.33  &  75.10  &  75.52  &  75.71  & 75.56  \\ \hline
   \end{tabular}}
   \caption{ \textbf{Impact of hyperparameter $\alpha$}. Both of temperature $\tau_f$ and $\tau_r$ are setting to 4.0. $\alpha=0.0$ means normal KD without reverse KL divergence, all the results are the average of 4 runs. }
   \label{tab:ablationalpha}
\end{table}

\begin{table}[htbp]
   \begin{center}
      \resizebox{0.5\textwidth}{!}{
         \begin{tabular}{c|c|c|c|c}
            \hline
            \multirow{2}{*}{$\tau_f$, $\tau_r$} & \multicolumn{3}{c}{ResNet32$\times$4 $\to$ ResNet8$\times$4} \\
            \cline{2-5}
            &  $2.0$, $8.0$  &  $4.0$, $4.0$  &  $8.0$, $2.0$ &  accumulate \\ \hline
            top-1             &  76.02  &  75.71  &  76.21   & 76.22 \\ 
            \hline
            \multirow{2}{*}{$\tau_f$, $\tau_r$} & \multicolumn{3}{c}{VGG13 $\to$ VGG8} \\
            \cline{2-5}
            &  $2.0$, $8.0$  &  $4.0$, $4.0$ & $8.0$, $2.0$  & accumulate \\ \hline
            top-1             &  74.74  &  74.23  &  73.50  &  74.69 \\ 
            \hline
         \end{tabular}
      }
      \caption{ \textbf{Impact of different temperature coefficients}. Hyperparameter $\alpha$ is setting to 4.0. All the results are the average of 4 runs. }
      \label{tab:ablationtemperature}
   \end{center}
\end{table}

\begin{figure}[htbp]
   \centering
   \begin{subfigure}[b]{0.45\textwidth}
      \centering
      \includegraphics[width=\textwidth]{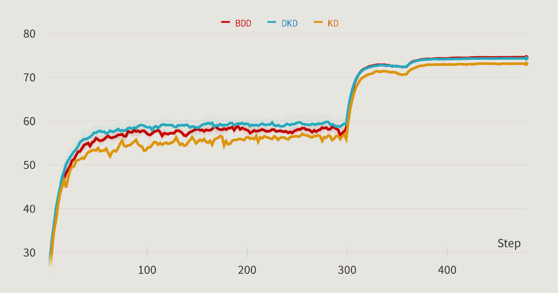}
      \caption{Top-1 accuracy($\%$) on validation.}
      \label{fig:fig3a}
   \end{subfigure}
   \begin{subfigure}[b]{0.45\textwidth}
      \centering
      \includegraphics[width=\textwidth]{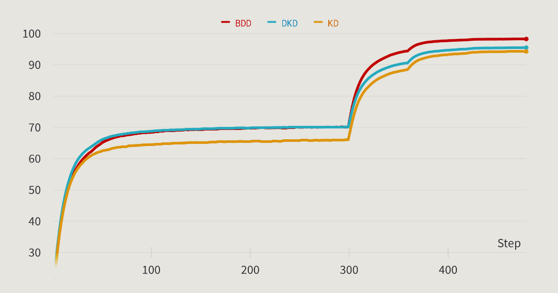}
      \caption{Top-1 accuracy($\%$) on train set.}
      \label{fig:fig3b}
   \end{subfigure}
   \caption{\textbf{Top-1 accuracy of training on validation and train sets}. The red line represents the accuracy of BDD, the blue line represents the accuracy of our DKD, and the yellow line represents the accuracy of KD.}
   \label{fig:fig3}
\end{figure}

We have also noticed intriguing phenomena during the training process. As depicted in figure(\ref{fig:fig3}), in the majority of training scenarios, BDD achieves top-1 accuracy levels on the validation set that match those of the state-of-the-art. Simultaneously, it manages to maintain higher accuracy on the training set(about $3\%-4\%$), while state-of-the-art methods often has little improve or even exhibit noticeable drops in training set accuracy. We believe that this characteristic not only indicates that there is still untapped potential in the student's representational capacity but also underscores the effectiveness of BDD.

\section{Conclusion}
In this paper, we propose an improved logit-based knowledge distillation method called Balance Divergence Distillation (BDD), which is a simple improvement of KL divergence. We introduce reverse KL divergence to balance the positive and negative samples in the knowledge distillation process. We also improve the BDD loss to further balance the positive and negative samples by using different temperature coefficients to regulate the enhancement and reduction of positive and negative aspects in BDD. We conduct experiments on both classification and semantic segmentation tasks to evaluate the performance of BDD. The results show that our method can achieve some improvement over the baseline student network. By adding the BDD loss to different attention outputs, state-of-the-art performance can be achieved. Compared with other knowledge distillation methods, our method achieves the best performance on both the same and different architectures. 
In the future, we will explore the application of BDD to other tasks, and we hope that this simple and effective method can be a good baseline for knowledge distillation to effectively train lightweight networks for other deep learning tasks.





{
    \bibliographystyle{elsarticle-num}
    \bibliography{elsarticle-template-num}
}

\end{document}